\documentclass[
]{ceurart}

\usepackage[frozencache=true,cachedir=minted-cache]{minted} 

\begin{document}

\copyrightyear{2021}
\copyrightclause{Copyright for this paper by its authors.
  Use permitted under Creative Commons License Attribution 4.0
  International (CC BY 4.0).}

\conference{CLEF 2021 -- Conference and Labs of the Evaluation Forum, 
	September 21--24, 2021, Bucharest, Romania}

\author{Ipek Baris Schlicht}[%
email=ibarsch@doctor.upv.es,
]

\author{Angel Felipe Magnossão de Paula}[%
email=adepau@doctor.upv.es,
]

\address{Universitat Politècnica de València, Spain}

\newif\ifproofread

\newcommand{\changemarker}[1]{%
\ifproofread
\textcolor{red}{#1}%
\else
#1%
\fi
}

\title{Unified and Multilingual \changemarker{Author} Profiling for Detecting Haters}
\title[mode=sub]{(Notebook for PAN at CLEF 2021)}



\begin{abstract}
\changemarker{This paper} presents a unified user profiling framework to identify hate speech spreaders by processing their tweets regardless of the language. The framework encodes the tweets with sentence transformers and applies an attention mechanism to select important tweets for learning user profiles. Furthermore, the attention layer helps to explain why a user is a hate speech spreader by producing attention weights at both token and post level. Our proposed model outperformed the state-of-the-art multilingual transformer models.

\end{abstract}

\maketitle              % typeset the header of the contribution

\section{Introduction}
\proofreadfalse
Hate speech is a type of online harm that expresses hostility toward individuals and social groups based on race, beliefs, sexual orientation, etc.~\cite{levy1986encyclopedia}. 
Hateful content \changemarker{is disseminated} faster and reaches \changemarker{wider} users than non-hateful contents through social media~\cite{MathewDG019,ziems}. 
This dissemination could trigger prejudices and violence. As a recent example of this, during the COVID-19 pandemic, people of Chinese origin suffered from discrimination and hate crimes~\cite{wang2021m,he2020discrimination}. 
Policymakers and social media companies work hard on mitigating hate speech and the other types of abusive language~\cite{preslav_survey_2021} while keeping balance of freedom of expression. 
AI systems are encouraged for easing the process and understanding the rationales behind hate speech dissemination~\cite{schmidt2017survey,FortunaN18}.

In \changemarker{natural language processing}, hate speech has been widely studied in social media (e.g~\cite{DBLP:conf/semeval/BasileBFNPPRS19,poletto2020resources}) or as a task of news comment moderation (e.g ~\cite{korencic-etal-2021-block,shekhar2020automating}). \changemarker{However, m}ajority of the prior studies formulates the problem as a text classification~\cite{macavaney2019hate,schmidt2017survey} that determines whether an individual post is hate speech. This year, PAN 2021 \changemarker{organization~\cite{bevendorff:2021b}} proposed to explore the task as an author \changemarker{profiling problem~\cite{rangel:2021}}. In this case, the objective is to identify possible hate speech spreaders on Twitter as an initial effort towards preventing hate speech from being propagated among online users~\cite{rangel:2021}.

\begin{figure*}
    \centering
    \includegraphics[width=\textwidth]{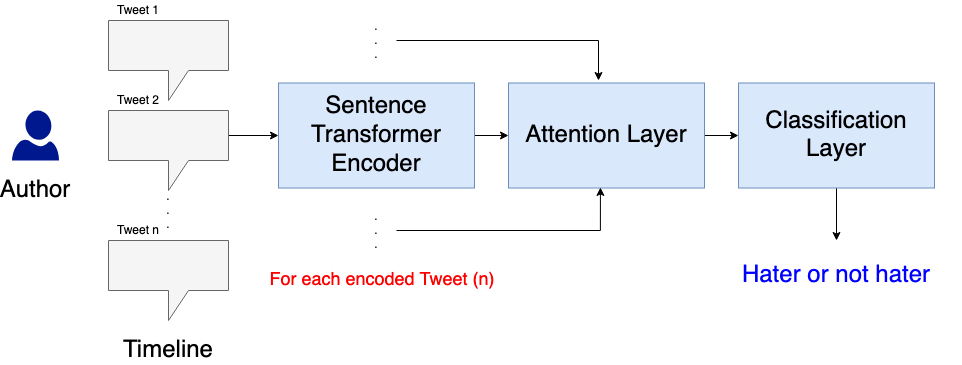}
    \caption{The Proposed Framework}
    \label{fig:system}
\end{figure*}

In a similar shared task on profiling fake news spreaders~\cite{PardoGGR20}, many approaches rely on appending tweets to one text for each user (e.g~\cite{vogel2020fake,buda2020ensemble,Pizarro20}) to encode the inputs. However, this approach could be problematic if not all the tweets shared by hate speech spreader\changemarker{s} convey hatred message\changemarker{s}, and a human moderator needs \changemarker{a} detailed justification to ban users or delete related tweets. Furthermore, the global issues such as COVID-19 attract heated discussions from the users worldwide, thus there is a need for supporting multi-language systems to moderate those discussions. With these motivations, we propose a unified framework which is scalable to other languages and explains why a user receives a certain label based on the language used in her tweets by using token level and post level attention mechanisms~\cite{VaswaniSPUJGKP17}\changemarker{,} as shown in Figure~\ref{fig:system}. Our model outperformed multilingual DistillBERT~\cite{DBLP:journals/corr/abs-1910-01108} models. The source code is publicly available\footnote{\url{https://github.com/isspek/Cross-Lingual-Cyberbullying}}.

\section{Methodology}
Our proposed framework is shown in Figure~\ref{fig:system}. The input of the framework is a \changemarker{author} profile \changemarker{that} posts \textbf{n} number tweets. Each post is encoded with a Sentence Transformer, and then the encoded tweets pass through an attention layer. Finally, the output of the attention layer is fed into a classification layer which decides whether the \changemarker{author} is a hate speech spreader \changemarker{or not}. We give more details of each component in the subsequent sections.

\subsection{Post Encodings}
We encode the tweets with a Sentence-BERT (SBERT)~\cite{ReimersG19}\changemarker{,} a modified BERT~\cite{DevlinCLT19} network \changemarker{and} consists of Siamese and Triplet network structures. SBERTs are computationally more effective than BERT models and could provide semantically more meaningful sentence representations. Like BERT models, SBERTs \changemarker{also have} variations~\cite{DBLP:journals/corr/abs-1910-03771} that are publicly available. Since we have a limited resource to train our framework, and aim to use \changemarker{a} language model that learn\changemarker{s} the usages of social language, we prefer the pre-trained SBERT that is trained on Quora corpus in 50 languages\changemarker{,} and its knowledge is distilled~\cite{reimers-2020-multilingual-sentence-bert}. The SBERT produces outputs with 768 hidden layers. We set \changemarker{the} maximum length of the post as 32, and apply zero padding on any texts shorter than 32 tokens. The sentence embeddings are obtained by mean pooling operation on the last hidden of the outputs.

\subsection{Post-Level Attention Layer}
We employ an attention layer in order to learn importance scores for determining \changemarker{author profile vectors}. 
First, the pooled tweets (\textit{Hp}) are projected by feeding them to \changemarker{a} linear layer which produces \changemarker{a} hidden representation of \changemarker{the author} profile (\textit{Hap}) as shown in  Equation~\ref{eq1}. 
Next, a softmax layer is applied to get the similarity between the post and \changemarker{author} profile \changemarker{(\textit{Hap})}. Lastly the similarity scores are multiplied with the \changemarker{author} profile to obtain \changemarker{the} attended \changemarker{author} profile (\textit{$Hap^{attended}$})\changemarker{,} as seen in Equation~\ref{eq2}. 

\begin{equation} \label{eq1}
Hap=HpWap+b^{T} \\
\end{equation}

\begin{equation} \label{eq2}
Hap^{attended} = softmax({Hp*Hap^{T}})Hap \\
\end{equation}

\subsection{Classification Layer}
The classification layer consists of two linear layers. The output of the first layer is activated with \changemarker{the} tanh function to learn the non-linearity in the features. The second layer outputs the probabilities for \changemarker{each class}. The input of the classification layer is the attended user profile followed by a dropout layer which prevents the over-fitting. We use a cross entropy loss function for the outputs of the classification layer and an Adam \changemarker{optimizer with a weight decay}. \changemarker{During} training, the weights of the models are optimized by minimizing the loss, \changemarker{and} the batches contain mixed English and Spanish samples. 

\section{Experiments}

\subsection{Dataset}
PAN \changemarker{Profiling} Hate Speech Spreader \changemarker{Task~\cite{rangel:2021}} contains \changemarker{a} dataset in English and Spanish, whose samples were collected from Twitter. The total number of the profiles are 200 for each language\changemarker{,} and each profile \changemarker{is composed of a feed of 200 tweets}. The class distribution of the dataset is highly balanced. We observe a significant difference between the length of \changemarker{tweets} by hate speech spreader\changemarker{s} and normal profiles in the Spanish set. The statistics of the dataset \changemarker{are} summarized in Table~\ref{tab:data_stats}.

\begin{table*}[]
    \caption{The statistics of the \changemarker{training} dataset}
    \centering
    \begin{tabular}{*{3}{l}r}
    \toprule
    \textbf{Stats} & \textbf{En} & \textbf{Es} \\
    \toprule
    \#Total Profiles & 200 & 200 \\
    \#Hate Speech Spreaders & 100 & 100 \\
    \#Tweets per Profile & 200 & 200\\
    \#Mean and Std of Tweets by hate speech Spreader& 67.72 $\pm$ 30.34 & 75.32 $\pm$ 28.91 \\
    \#Mean and Std of Tweets by Normal Profiles& 67.42 $\pm$ 29.05 & 68.47 $\pm$ 28.99\\
    \bottomrule
    \end{tabular}
    
    \label{tab:data_stats}
\end{table*}

\subsection{Preprocessing}
The \changemarker{organizers have already cleaned the samples in the dataset}. For example, certain patterns have been replaced with special tags. We extend the vocabulary of the models' tokenizers with these tags as follows:
\begin{itemize}
    \item \texttt{\#URL\#} is replaced with \texttt{[URL]}
    \item \texttt{\#HASHTAG\#} is replaced with \texttt{[HASHTAG]}
    \item \texttt{\#USER\#} is replaced with \texttt{[USER]}
    \item \texttt{RT} is replaced with \texttt{[RT]}
    
\end{itemize}

\subsection{Baselines and Ablation Models}
We compare the performance of our model with a set of baselines and an ablation model as follows:
\begin{itemize}
    \item \textbf{DistillBERT}~\cite{DBLP:journals/corr/abs-1910-01108}: We use one of its version that is multilingual and cased sensitive. First each tweets of an \changemarker{author} is joined to obtain one text. Then the joined texts for each users are fine-tuned with the DistillBERT by keeping their maximum length as 500 tokens.
    \item \textbf{DistillBERT{*}}: We additionally add \texttt{[POSTSTART]} and \texttt{[POSTEND]} tags, which \changemarker{indicate} the start and the end of the \changemarker{tweets}, to the vocabulary of the extended DistillBERT tokenizer. 
    \item \textbf{SBERT-Mean}: is an ablation model that replaces the attention layer with a mean pooling layer which computes the mean values of the \changemarker{tweets'} hidden representations. 
    
\end{itemize}

\subsection{Training Settings}
We train the models by applying 5-Fold Cross Validation\footnote{We experiment also 10-Fold, but the models show worse performance in the test set.}, with the epochs of 5, learning rate as 1e-5, batch size as 2. We use the GPU of the Google Colab\footnote{\url{https://colab.research.google.com/}} as \changemarker{an} environment \changemarker{for training the models}. We use a fixed random seed of 1234 to ensure reproducible results. \changemarker{The official results are obtained by a TIRA machine~\cite{potthast:2019n}.}

\section{Results and Discussion}
We report the F1-Macro, F1-Weighted, accuracy, precision\changemarker{,} and recall for each model. 
Table~\ref{tab:5_fold} presents the results of the 5-fold cross validation training. SBERT-Attn, the model that we propose, outperformed the other models in all metrics. When we compare SBERT-Mean and SBERT-Attn, we see that standard deviations of the SBERT-Attn \changemarker{are} lower than the ablation model. This result indicates that the attention layer enables more generalized feature representations\changemarker{. It} also shows that the tweets by the hate speech spreader \changemarker{are} not necessarily hatred \changemarker{tweets} and vice versa for the \changemarker{non haters}. For this reason, the DistillBERT models that joins the all tweets by the user to one underperformed.
\begin{table*}[!ht]
\centering
 \caption{The results of the 5 Fold Cross Validation Experiment}
 \begin{tabular}{*{7}{l}r}
 \toprule
 \textbf{Models} & \textbf{F1-Macro} & \textbf{F1-Weighted} & \textbf{Accuracy} & \textbf{Precision} & \textbf{Recall }\\
 \toprule
\textbf{DistillBERT} & 67.46 $\pm$ 5.28 & 67.58 $\pm$ 5.37 & 67.75 $\pm$ 5.15 & 67.04 $\pm$ 5.68 & 71.46 $\pm$ 1.63 \\
\textbf{DistillBERT*} & 61.90 $\pm$ 3.01 & 62.04  $\pm$ 3.22 & 62.25 $\pm$ 3.39 & 63.13 $\pm$ 4.40 & 59.86 $\pm$ 7.49 \\
 \midrule
\textbf{SBERT-Mean} & 69.55  $\pm$  6.82 & 69.58 $\pm$ 6.71 & 69.75 $\pm$ 6.86 & 67.38 $\pm$ 3.61 & 77.10 $\pm$ 12.12 \\
\textbf{SBERT-Attn} & \textbf{73.62}  $\pm$ 4.11 & \textbf{73.77} $\pm$ 4.12 & \textbf{74.0} $\pm$ 4.14 & \textbf{70.97} $\pm$ 5.39 & \textbf{81.23}  $\pm$ 5.39 \\
 \bottomrule
 \end{tabular}
 \label{tab:5_fold}
\end{table*}

\begin{table*}[!ht]
    \centering
    \caption{Cross validation for each language and the PAN shared official result.}
    \begin{tabular}{*{3}{l}r}
        \toprule
         \textbf{Mode}& \textbf{Language} & \textbf{Accuracy} \\
         \toprule
         Cross-Val & En & 67.09 $\pm$ 7.88 \\
         & Es & 80.54  $\pm$ 1.78 \\
         \midrule
         Official Result & En & 58 \\
         & Es & 77 \\
         \bottomrule
    \end{tabular}

    \label{tab:off_result}
\end{table*}

For the submission to the PAN shared task, we leverage the 5-fold trained models to obtain the predictions on official test set. The final predictions are the majority class. Table~\ref{tab:off_result} shows cross validation results for the English samples and the Spanish samples, and the official results of the PAN shared task where the accuracy is the evaluation metric. Our model obtained a result \changemarker{with} similar range in cross-validation. The performance of the English set is worse than the Spanish \changemarker{one}. Cultural bias or the topical difference could be reasons \changemarker{for} the performance. We leave the detail\changemarker{ed} analysis of these issues as future work.

\section{Visualizations}
\begin{figure}[!ht]
    \centering
    \includegraphics[width=\textwidth]{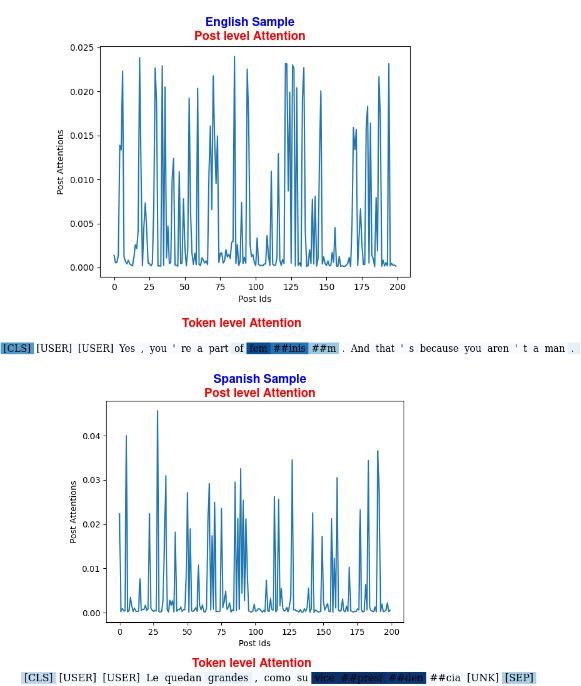}
    \caption{Attention visualizations for English and Spanish. The original sentence in English is \changemarker{[USER] [USER] \textit{Yes, you're a part of feminism. And that's because you aren't a man};} and the other in Spanish is \changemarker{[USER] [USER] \textit{Le quedan grandes, como su vicepresidencia}} (Some emojis)}
    \label{fig:visualization}
\end{figure}
Our framework can provide explanations with \changemarker{tweet-level} and token-level attention, as shown in Figure~\ref{fig:visualization}. The token-level attentions are the average of the attentions in the last layer of the SBERT \changemarker{and} they are obtained through the self-attention mechanism. The \changemarker{tweet-}level attentions are obtained with the attention layer, which is connected to the classification layer. The examples in the figure are the most hatred examples from the \changemarker{authors} that are analysed. In \changemarker{the} English example, the model pays attention to feminism. In \changemarker{the} Spanish example, \textit{vice presidencia} is the important entity.

\section{Conclusion}
In this paper, we presented a unified framework for monitoring \changemarker{hate speech spreaders} in \changemarker{multilingualism}. The framework leverages \changemarker{multilingual} SBERT representations to encode texts regardless of \changemarker{the} language \changemarker{and} uses \changemarker{an} attention mechanism to determine the importance of the tweets by the \changemarker{author in} the task. Our method\changemarker{s} outperformed multilingual DistillBERT and SBERT that \changemarker{apply} mean pooling on the \changemarker{tweets.}

In the future, we plan to evaluate the method on the related user profiling tasks such as \changemarker{profiling} fake news spreaders~\cite{PardoGGR20} and investigate advanced method (e.g~\cite{pfeiffer2020adapterhub}) for effectively transferring knowledge across the languages.

% \section{Appendix}
% We experiment also 10-Fold cross validation. The results are shown in Table~\ref{tab:10_fold}. SBERT-Attn is the best model also in this version. However, the predictions that is obtained by 10-fold training, has shown worse performance on the official test set. The model got the accuracy 57\% on the English samples and 74\% on the Spanish ones.

% \begin{table}[]
%     \centering
%  \begin{tabular}{*{7}{l}r}
%  \toprule
%  \textbf{Models} & \textbf{F1-Macro} & \textbf{F1-Weighted} & \textbf{Accuracy} & \textbf{Precision} & \textbf{Recall }\\
%  \toprule
% \textbf{DistillBERT} & 64.83 $\pm$ 4.23 & 64.94 $\pm$ 4.44 & 65.5 $\pm$ 4.0 & 65.16  $\pm$ 6.47 & 69.59 $\pm$ 12.94 \\
% \textbf{DistillBERT*} & 60.16 $\pm$ 5.09 & 60.5  $\pm$ 5.30 & 60.5 $\pm$ 5.34 & 60.52 $\pm$ 7.62 & 62.38 $\pm$ 7.92 \\
%  \midrule
% \textbf{SBERT-Mean} & 70.52 $\pm$ 8.32 & 70.68 $\pm$ 8.30 & 71.00 $\pm$ 8.23 & 70.84 $\pm$ 11.55 & 74.49 $\pm$ 11.16 \\
% \textbf{SBERT-Attn} & 71.72 $\pm$ 8.22 & 71.73 $\pm$ 8.37 & 72.5 $\pm$ 7.75 & 68.20 $\pm$ 8.99 & 86.77 $\pm$ 8.94 \\
%  \bottomrule
%  \end{tabular}
%     \caption{10-Fold Cross Validation Results}
%     \label{tab:10_fold}
% \end{table}

\bibliography{paper}

\end{document}